\documentclass{article}
\PassOptionsToPackage{numbers, compress}{natbib}
\usepackage{array}
\usepackage{wrapfig}

\usepackage[preprint]{neurips_2026}

\usepackage[utf8]{inputenc} 
\usepackage[T1]{fontenc}    
\usepackage{hyperref}       
\usepackage{url}            
\usepackage{booktabs}       
\usepackage{amsfonts}       
\usepackage{nicefrac}       
\usepackage{microtype}      
\usepackage{xcolor}         
\usepackage{colortbl}

\usepackage{amsmath}
\usepackage{amssymb}
\usepackage{mathtools}
\usepackage{amsthm}
\usepackage{multirow}
\usepackage{capt-of}
\usepackage{wrapfig}
\usepackage{graphicx}

\usepackage[capitalize,noabbrev]{cleveref}

\theoremstyle{plain}

\theoremstyle{definition}

\theoremstyle{remark}

\title{SlimVLM: Sensitivity-aware Dynamic Structured Pruning with Adaptive Visual Token Selection for Efficient Vision-Language Models}

\author{%
  Yaozhi Wen\textsuperscript{1}\thanks{Equal contribution.} \quad
  Jialong Guo\textsuperscript{1}\footnotemark[1] \quad
  Zhenliang Ni\textsuperscript{1}\footnotemark[1] \quad
  Han Shu\textsuperscript{1} \quad
  Xinghao Chen\textsuperscript{1}\footnotemark[2]\thanks{Corresponding author: xinghao.chen@huawei.com} \\
  \textsuperscript{1}Huawei Technologies \\
}

\begin{document}

\maketitle

\begin{abstract}
  While Vision-Language Models (VLMs) have demonstrated remarkable performance in processing and understanding both text and images, their large parameter sizes lead to significant computational overhead, limiting their deployment on resource-constrained devices. While pruning has been effective for compressing Large Language Models (LLMs), directly applying it to VLMs leads to significant performance drops, largely due to redundant visual tokens interfering with importance estimation. To this end, we propose \textbf{SlimVLM}, a structured pruning framework designed to compress VLMs while preserving their task performance. We introduce an adaptive visual token selection strategy for VLMs that leverages average text-to-visual attention scores to assess the importance of visual tokens, removing redundant ones during pruning based on a set threshold, thereby optimizing the importance calculation. Recognizing the varying tolerance to sparsity across different modules, we also propose a Sensitivity-aware dynamic pruning mechanism that determines the appropriate pruning ratio for each module by calculating the linear reconstruction error between the outputs of the pruned and unpruned modules, ensuring overall performance stability. Experimental results show that SlimVLM outperforms existing methods across multiple multimodal benchmarks, achieving state-of-the-art performance.
\end{abstract}

\section{Introduction}

By extending large language models (LLMs) to incorporate visual information, Vision-Language Models (VLMs)~\cite{llava,qwen2_5,internVL3_5} have emerged as a pivotal area of research in visual question answering~\cite{guo2023imagestextualpromptszeroshot,Yu_2025}, embodied task planning~\cite{mendezmendez2023embodiedlifelonglearningtask} and dialogue assistants~\cite{li2018endtoendtaskcompletionneuraldialogue}. While VLMs demonstrate impressive performance on tasks such as visual question answering~\cite{VQA} and multimodal reasoning~\cite{mmmu}, their substantial parameter sizes introduce significant computational overhead during inference, limiting their deployment on resource-constrained devices like smartphones and autonomous driving systems. A critical challenge, therefore, lies in reducing model parameters while maintaining performance.

In VLMs, visual features extracted by the image encoder are typically projected into the language model's input space via a projector, a process that introduces a large number of visual tokens. For instance, LLaVA-1.5~\cite{llava} employs 576 visual tokens, while Qwen2.5-VL~\cite{qwen2_5} generates even more at its native resolution. In contrast, text tokens constitute only a small fraction of the input. However, not all visual tokens contribute meaningfully to the task. For example, background regions often receive lower attention weights, indicating significant redundancy~\cite{VisionZip}. As a result, current research on VLM compression has primarily focused on pruning 
\begin{wrapfigure}{r}{0.56\linewidth} 
  \centering
  \includegraphics[width=\linewidth]{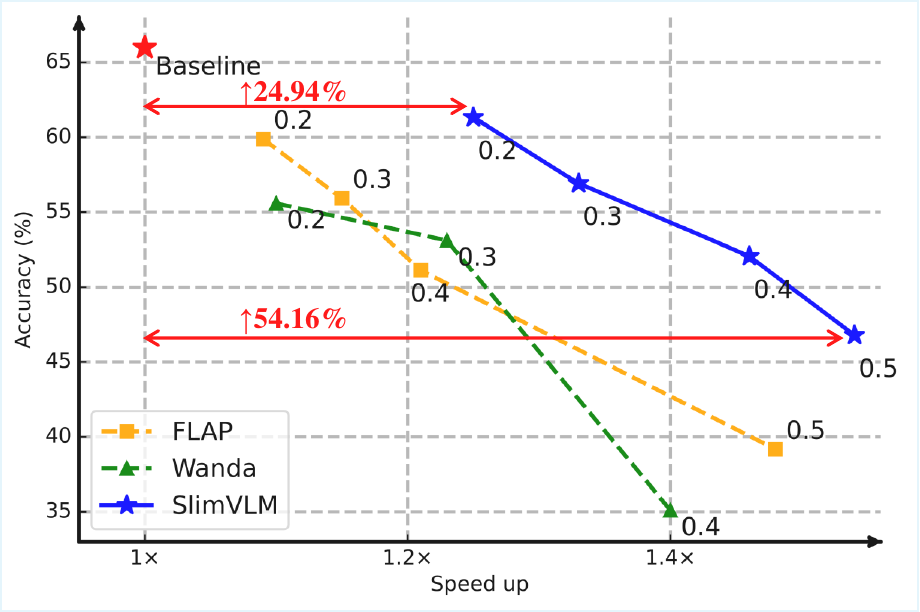}
  \caption{The trade-off between accuracy and inference speed on ScienceQA with SlimVLM.}
  \vspace{-10pt}
  \label{fig3}
\end{wrapfigure}
these redundant visual tokens to accelerate inference~\cite{fastv,HiPrune}. While these methods improve computational efficiency, their ability to reduce model parameters remains limited. An alternative approach is to prune the language model within VLMs; however, the presence of a large number of visual tokens not only complicates module-importance calculation but also introduces interference from irrelevant visual tokens, which can distort importance estimation and lead to suboptimal pruning decisions. As a result, directly applying existing pruning methods to the language model of a VLM leads to significant performance degradation.

Considering the aforementioned issues, we propose SlimVLM, a structured pruning framework tailored for Vision-Language Models. Prior to pruning, we introduce a VLM-specific visual token selection strategy to identify and retain the most informative visual tokens, thereby mitigating the interference from irrelevant ones during module importance estimation. Our observation indicates that visual tokens receiving higher attention typically convey richer semantic information and are more critical for accurate structural importance assessment. Specifically, we leverage the attention maps from decoder layers and compute the importance scores of visual tokens using text queries. Low-scoring tokens are filtered out, and the resulting token mask matrix is then applied to calculate module importance for subsequent pruning.

Furthermore, we introduce a Sensitivity-aware pruning ratio adjustment mechanism to minimize output deviations across different modules. This problem stems from divergent sparsity tolerance among modules, where crucial ones demand higher parameter density to maintain functionality. Specifically, we employ the Pearson correlation coefficient between module outputs before and after pruning as a linear metric and dynamically adjusts the pruning ratio for each module, achieving global adaptive optimization and forming a precise and efficient adaptive dynamic pruning strategy. During structured pruning, we remove both attention heads and MLP channels. For models adopting a Multi-Head Attention structure, each attention head is treated as an independent unit, and the combination that exhibits the highest Pearson correlation with the original output is retained. We further adapt this strategy to Grouped-Query Attention to ensure robustness and generality across different attention mechanisms. When computing channel importance, we redefine the structural importance criterion in the output feature space, while considering both the magnitude and directional consistency of weights.

We conduct extensive experiments to validate the effectiveness of the proposed approach. Compared with existing structured pruning methods, our proposed method achieves superior performance under the same pruning ratio, retaining near-original accuracy under 20\% pruning, and reaches state-of-the-art results. As shown in Figure \ref{fig3}, our method demonstrates significant advantages in inference speed and accuracy compared to other methods at the same pruning ratio. Moreover, the proposed visual token selection mechanism can be seamlessly integrated into other pruning frameworks, significantly improving their pruning efficiency and overall performance. Our contributions can be summarized as follows:
\begin{itemize}
	\item We propose a structured pruning framework named SlimVLM, and introduce a VLM-specific visual token selection strategy that leverages text-to-visual attention to filter eligible tokens for pruning.
    \item We introduce a Sensitivity-aware dynamic pruning strategy that adapts pruning rates across different modules to maintain global performance stability while achieving a more balanced compression.
    \item Extensive experiments demonstrate that SlimVLM achieves competitive accuracy while pruning 20\% of the LLaVA-1.5 model, outperforming existing structured pruning methods in terms of stability and performance.
\end{itemize}

\section{Related Work}

\textbf{Vision-language Models.} Enabled by the open-sourcing of LLMs like LLaMA~\cite{llama} and Qwen~\cite{bai2023qwentechnicalreport}, VLMs can seamlessly integrate textual and visual information through modality alignment, enhancing the model's understanding of multimodal data. The pre-trained language model serves as the backbone for text generation, processing multimodal inputs in an autoregressive manner to produce coherent and contextually relevant outputs~\cite{Yin_2024}. Recent work on vision-language models has shown that expanding the parameter size of foundational language models can significantly improve multimodal understanding and generation~\cite{qwen2_5,internVL3_5}. However, the massive parameter sizes pose substantial challenges for practical deployment. Furthermore, the use of higher-resolution images inevitably leads to an exponential increase in the visual sequence length, with these visual tokens often being sparse in information~\cite{ZipVL,DivPrune}.

\textbf{Advances in pruning methods.} Pruning is a model compression technique that identifies and removes redundancy in the structure or parameters of a neural network~\cite{Puzzle,he2024matterstransformersattentionneeded,sengupta2025pruneoncedesigningcalibrationfree}. For language models, the importance of components within the network is evaluated using a small-scale validation dataset, allowing for the safe removal of redundant components. DLP~\cite{DLP} achieves sparsity by removing individual weight connections within the network; however, this unstructured sparsity is not optimal for hardware acceleration. In contrast, structured pruning is more favorable. ShortGPT~\cite{ShortGPT} achieves sparsity by removing unimportant decoder layers, while SlimLLM~\cite{slimllm} achieves structured sparsity by removing attention heads and channels. For vision-language models, a range of token pruning methods have been proposed to accelerate inference~\cite{VisionThink,LLaVA-PruMerge,VScan,liu2024multistagevisiontokendropping}. The most common approach is to exploit the sparsity of attention within multimodal data to guide compression. Tokens with low attention scores can often be removed without significantly affecting the original computation. Attention-based compression strategies can generally be categorized into those applied within the encoder~\cite{LLaVA-PruMerge,VisionZip} and those within the decoder~\cite{ye2024fitprunefasttrainingfree,lin2025boostingmultimodallargelanguage}. However, this token pruning approach does not effectively reduce the model parameter size, limiting its deployment on smaller devices. Additionally, the introduction of a large number of visual tokens makes it difficult to directly apply pruning methods designed for language models to vision-language models. By introducing an adaptive token selection strategy prior to module importance calculation, we enhance the compatibility of existing pruning methods with VLM compression.

\section{Method}

\begin{figure*}[t]
  \centering
  \includegraphics[width=\linewidth]{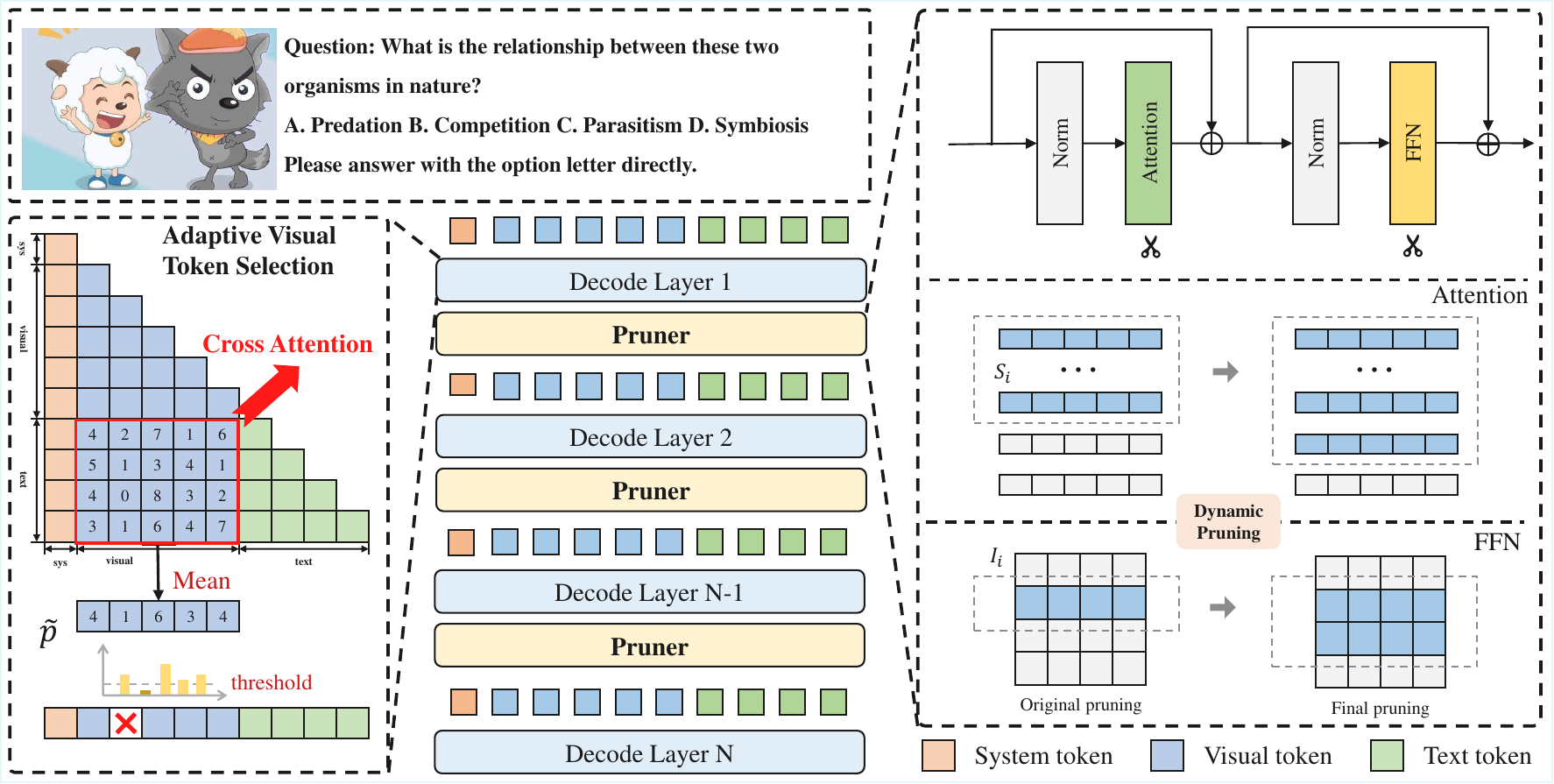}

   \caption{\textbf{The overall framework of our proposed SlimVLM}. Our approach consists of two main components: (1) Adaptive Visual Token Selection, which reuses decoder attention maps and text queries to compute importance scores for filtering out less informative visual tokens; and (2) Structured Pruning, which removes redundant attention heads and MLP channels based on their importance scores $S_i$ and $I_i$, respectively.}
   \label{fig1}
\end{figure*}

For pruning the language model within Vision-Language model, we argue that particular attention should be given to the variation in input modalities, which encompasses not only textual tokens but also a significant proportion of visual tokens. Furthermore, we introduce a dynamic pruning strategy that adaptively adjusts pruning ratios based on the linear approximation error of outputs before and after pruning attention heads and MLP layers. The overall architecture is shown in Figure \ref{fig1}.

\subsection{Adaptive Visual Token Selection}

The extensive redundant visual tokens in VLMs pose a significant challenge to structural pruning, as they interfere with the accurate assessment of module importance, which is essential for determining safe removal. To avoid this issue, we leverage the inherent sparsity of visual feature attention to select appropriate visual tokens before the pruning stage. Specifically, for each decoder layer involved in pruning, we independently compute a visual-token mask using the text-to-visual attention map from that layer. This mask is used only for module-importance estimation and is not applied during the final inference of the pruned model.

VLM decoder layers adopts the causal attention mechanism from the transformer architecture~\cite{transformer}. Without loss of generality, we describe the single-head attention below. Formally, the attention matrix $A \in \mathbb{R}^{L \times L}$, where $L$ denote the length of all input tokens (including both visual tokens and text tokens), is computed by:
\begin{equation}
  A = \text{Attention}(Q,K) = \text{Softmax}(\frac{QK^T}{\sqrt{D}}),
  \label{eq:important}
\end{equation}
where $D$ denote the matrix dimension, $Q \in \mathbb{R}^{L \times D}$ and $K \in \mathbb{R}^{L \times D}$ are the query matrix and key matrix, respectively.

As the model may focus on different parts of the image when handling different prompts, we ground our token selection process in the cross-modal interplay between the text-query dimension and the visual-key dimension. In particular, we extract the cross-attention matrix $P \in \mathbb{R}^{L_t \times L_v }$ from attention matrix A, $L_t$ and $L_v$ are the lengths of text and visual tokens: 
\begin{equation}
    P = \left[ A_{i,j} \right], \text{ and } (i,j) \in \{\mathbb{T},\mathbb{I}\},
\end{equation}
where $\mathbb{T}$ and $\mathbb{I}$ denote the index sets of text prompt tokens and image tokens, respectively. 

Next, to suppress prompt-specific fluctuations and obtain a stable, prompt-agnostic importance ranking, we calculate the average attention score $p$ of each text-query pair for visual-key, which serves as the selection metric:
\begin{equation}
  p = [ p_1, p_2, \cdots , p_{L_v}] = \frac{1}{L_t}\sum^{L_t}_{i=1} P_i,
\end{equation}
where $P_i$ is the set of visual attention score corresponding to the $i$-th text query.

Previous research has found that in VLMs, visual tokens contain dense information in shallow layers and gradually become sparse in deeper layers, while text tokens maintain high attention consistently~\cite{pyramiddrop}. Therefore, during pruning, we retain all text tokens and filter out the visual tokens whose average attention score falls below the threshold $\alpha = \text{ mean }(p) * \beta$, where $\text{ mean }()$ denotes the averaging function, and $\beta$ is a hyperparameter. We set $\beta = 0.2 $ to ensure that the number of retained visual tokens is close to that of text tokens, and then obtain the selection mask $M=\{M^T,M^V\}$. Here,$M^T$ denotes the text mask, where all elements are set to True; and $M^V$ denotes the visual mask, which is given by the formula below:
\begin{equation}
    M_i^V = \left\{
    \begin{aligned}
    & \text { False } , \text {if } p_i < \alpha,\\
    & \text { True } , \text { otherwise },\\
    \end{aligned}
    \right. 
\end{equation}
where False indicates that the token is discarded, and True indicates that it is retained.

\subsection{Sensitivity-aware Dynamic Pruning Ratio }

Since the sparsity tolerance of different modules in a model typically varies, applying a uniform pruning strategy across all modules is generally suboptimal. In our work, we determine the appropriate pruning ratio for each module based on the error in linearly reconstructing its output from that of the unpruned module. We adopt Pearson correlation as a metric to quantify the degree of linear dependence, which is defined by the following formulation:

\begin{equation}
	L = Pearson(O_{pruned}, O_{raw}),
\end{equation}

where $O_{pruned}$ is the output of MHA or MLP after pruning, $O_{raw}$ is the output of MHA or MLP without pruning. A high degree of linear correlation implies a lower linear reconstruction error, suggesting that the corresponding module can tolerate a higher pruning ratio. The relationship between the pruning ratio and linear correlation is formally defined as follows:
\begin{equation}
\label{eq:9}
	r = r_{0}e^{\gamma(L/T - 1)},
\end{equation}
where $r_{0}$ denotes the initial pruning ratio, $\gamma$ is a hyperparameter controlling the sensitivity of adjustment, and $T$ represents the expected threshold for linear correlation. The actual pruning ratio is adaptively scaled: it increases when the measured linear correlation surpasses $T$, and decreases when it falls below $T$, thereby enabling a dynamic and module-specific pruning strategy.

Finally, we reduce the error caused by pruning through the method of least squares. The objective function is given as follows:
\begin{equation}
	J = min_{\eta, \zeta}||(\eta O_{pruned} + \zeta) - O_{raw}||_{2},
\end{equation}
After $\eta$ and $\zeta$ are computed, both of them can be fused with the output linear layer of MLP or MHA.

\subsection{Structured Pruning for Decoder}

In this subsection, we present the pruning method in SlimVLM, which simultaneously prunes attention heads and MLP channels.

\textbf{Heads Pruning}. In the pruning of multi-head attention (MHA) heads, we aim to preserve the subset of heads whose combined output best approximates the original MHA module output in terms of linear correlation. To identify such heads, we first compute the similarity between the output of the MHA module with each individual head removed and the  unpruned MHA output, as defined by the following equation:
\begin{equation}
	S_{i} = Pearson(O^{MHA}, O^{MHA} - O^{MHA}_{i}),
\end{equation}
where $O^{MHA}$ represents the output of MHA, $O^{MHA}_{i}$ is the $i$-th head's output. We employ Pearson correlation to measure the linear correlation. A higher $S_{i}$ indicates a smaller impact of removing the corresponding head on the output. Given the target pruning ratio, we initialize the candidate set with the heads exhibiting the lowest scores and employ an iterative optimization strategy, whereby pruned heads are sequentially substituted, to search for the optimal head configuration that maximizes output similarity.

Furthermore, when the Grouped-Query Attention architecture is applied in LLMs, our pruning strategy is specifically applied to the query heads. Notably, we do not impose the constraint of maintaining an equal number of heads across all groups, allowing for heterogeneous head distribution within groups post-pruning.

\textbf{Channels Pruning}. We assess the importance of intermediate channels based on activations and the down matrix $W^{down} \in \mathbb{R}^{D \times M}$. First, we perform principal component analysis on the MLP output $O^{MLP} \in \mathbb{R}^{N \times D}$ to obtain the corresponding eigenvalues $E \in \mathbb{R}^{D}$ and eigenvectors $W^{\prime} \in \mathbb{R}^{D \times D}$. The importance score of the $i$-th channel can be calculated as follows:
\begin{equation}
	C = Sigmoid(E / \bar{E}),
\end{equation}
\begin{equation}
	I_{i} = ||C_{0}W^{\prime}_{0}W^{down}_{i},...,C_{D}W^{\prime}_{D}W^{down}_{i}||_{2} \cdot ||X_{i}||_{2},
\end{equation}
where $\bar{E}$ denotes the mean of eigenvalues $E$, $W^{down}_{i}$ represents the $i$-th col of $W^{down}$, $X_{i}$ is the input of $i$-th channel. According to the importance, channels with lower scores will be pruned.

\section{Experiments}
\subsection{Setup}

\textbf{Models.} We conduct experiments on VLMs with different architectures and input resolutions to validate the effectiveness of the proposed SlimVLM. Our primary focus is on LLaVA-1.5-7B~\cite{llava} and LLaVA-Next-7B~\cite{llavanext}.
LLaVA-1.5 is the most widely used open-source VLM model, with the language model being Vicuna-7B, capable of handling up to 576 visual tokens. LLaVA-Next is the high-resolution extension of LLaVA-1.5, supporting up to 2880 visual tokens and demonstrating superior performance with high-resolution images. To evaluate SlimVLM's compatibility with the Grouped-Query Attention architecture, we also perform experiments on Qwen2.5-VL~\cite{qwen2_5}.

\textbf{Benchmarks.} For image-based multimodal evaluation, we conduct experiments on eight widely adopted benchmarks, including GQA dataset~\cite{GQA}, MMBench~\cite{MMBench} (which includes both EN and CN versions), MME~\cite{MME}, POPE~\cite{POPE}, SQA~\cite{SQA}, VQA-v2~\cite{VQAv2}, TextVQA~\cite{TextVQA}, and VizWiz~\cite{VizWiz}.

\textbf{Baselines.} We compare SlimVLM against two prior pruning methods: \textbf{FLAP}~\cite{flap} and \textbf{Wanda}~\cite{wanda}. We adapt Wanda to structured pruning by using the L2-norm of an entire weight group as the group's importance score. We refer to this structured variant as \textbf{Wanda-sp}.

\textbf{Implementation details.} The module importance was computed on a development set consisting of 24 randomly sampled instances from both the English and Chinese versions of MMBench. During the pruning process, we skipped several of the most important layers to preserve model functionality. Specifically, for the LLaVA series, layers 4 to 30 were pruned, while for Qwen2.5-VL, layers 2 to 27 were pruned. All pruning experiments were conducted on a single 32G GPU. For the 20\% pruning setup, the hyperparameters in Equation (\ref{eq:9}) were set to $\gamma = 5$ and $T = 1$. When the pruning ratio was increased to 40\%, the value of $\gamma$ was reduced to 1 to account for the increased $L$ after pruning. All evaluations were carried out using the lmms-eval library \cite{lmms_eval2024} across 8 32G GPUs. Due to hardware constraints, we employed lmms-eval-lite to reduce memory usage during the evaluation of VQA-v2, and set max pixels to $1280 \times 28 \times 28$ for the Qwen2.5-VL.

\renewcommand{\arraystretch}{1.2} 
\setlength\heavyrulewidth{0.25ex}
\begin{table*}[tb!]
\centering
\caption{\textbf{Performance of SlimVLM on LLaVA-1.5 and LLaVA-Next across Multiple Multimodal Benchmarks}. The “Average” metric is computed as the mean relative performance across datasets, normalized by the corresponding vanilla model performance. The \textbf{"Bolded"} represents the best result under the same pruning ratio, and "\underline{underline}" denotes the second best.}
\resizebox{\textwidth}{!}{%
\begin{tabular}{c|c|c|ccccccccc|c}
\toprule
 & Ratio & Method & GQA  & MMB & MMB$^{\text{CN}}$ & MME & POPE & SQA$^{\text{I}}$ & VQA$^{\text{2}}$ & VQA$^{\text{T}}$ & VizWiz & Average(\%)  \\ 
 \midrule \multirow{8}{*}{\rotatebox[origin=c]{90}{\textcolor{black}{\textbf{LLaVA 1.5-7B}}}} %

& 0\% & \cellcolor{gray!20}Vanilla
& \cellcolor{gray!20}61.94 & \cellcolor{gray!20}62.02
& \cellcolor{gray!20}50.26 & \cellcolor{gray!20}1780.64
& \cellcolor{gray!20}85.89 & \cellcolor{gray!20}65.99
& \cellcolor{gray!20}69.56 & \cellcolor{gray!20}48.07
& \cellcolor{gray!20}54.48 & \cellcolor{gray!20}100 \\ \cline{2-13}

& \multirow{3}{*}{20\%}  & Wanda-sp & \underline{57.19} & 51.72& 40.64 & 1406.53 & 84.12& 55.58 & \textbf{66.62}& \underline{41.20} & 37.34 & 85.31 \\
&  & FLAP & 55.45 & \underline{52.58} & \underline{41.24} & \underline{1645.92} & \underline{84.59} & \underline{59.89} & 64.32 & 40.59 & \underline{48.72} & \underline{89.37} \\

& & \cellcolor{blue!10}SlimVLM
& \cellcolor{blue!10}\textbf{58.00} & \cellcolor{blue!10}\textbf{56.44}
& \cellcolor{blue!10}\textbf{45.88} & \cellcolor{blue!10}\textbf{1682.96}
& \cellcolor{blue!10}\textbf{85.49} & \cellcolor{blue!10}\textbf{61.33}
& \cellcolor{blue!10}\underline{65.90} & \cellcolor{blue!10}\textbf{41.56}
& \cellcolor{blue!10}\textbf{54.55} & \cellcolor{blue!10}\textbf{93.80} \\

\cline{2-13}
& \multirow{3}{*}{40\%} & Wanda-sp & \underline{42.77} & 23.63 & 14.66 & \underline{1215.90} & \underline{82.39} & 35.1 & \underline{53.24} & \textbf{30.73} & 26.17 & \underline{60.25} \\
&  & FLAP & 42.11 & \underline{35.74} & \underline{20.70} & 796.17 & 64.31 & \underline{51.12} & 45.32 & 21.87 & \underline{32.35} & 59.32 \\

& & \cellcolor{blue!10}SlimVLM
& \cellcolor{blue!10}\textbf{46.44} & \cellcolor{blue!10}\textbf{43.57}
& \cellcolor{blue!10}\textbf{33.81} & \cellcolor{blue!10}\textbf{1331.00}
& \cellcolor{blue!10}\textbf{83.71} & \cellcolor{blue!10}\textbf{52.06}
& \cellcolor{blue!10}\textbf{54.10} & \cellcolor{blue!10}\underline{25.31}
& \cellcolor{blue!10}\textbf{44.89} & \cellcolor{blue!10}\textbf{75.16} \\

\midrule 

\multirow{8}{*}{\rotatebox[origin=c]{90}{\textcolor{black}{\textbf{LLaVA-Next-7B}}}} 

& 0\% & Vanilla & 63.67 & 66.15 & 55.33 & 1783.04 & 87.04 & 69.36 & 75.36 & 64.09 & 58.78 & 100  \\ \cline{2-13}

& \multirow{3}{*}{20\%}  & Wanda-sp & 56.97 & \underline{55.15} & 27.32& 1544.80&\underline{85.74} &57.56 &66.92 & \underline{57.46}&53.66 &84.46 \\
&  & FLAP & \underline{58.28}& 54.64& \underline{39.86}& \underline{1630.89}& 85.36 & \underline{59.44}& \underline{70.46}& 57.20 & \textbf{61.92}& \underline{89.95}\\

& & \cellcolor{blue!10}SlimVLM
& \cellcolor{blue!10}\textbf{61.22} & \cellcolor{blue!10}\textbf{59.52}
& \cellcolor{blue!10}\textbf{47.28} & \cellcolor{blue!10}\textbf{1725.55}
& \cellcolor{blue!10}\textbf{85.87} & \cellcolor{blue!10}\textbf{63.47}
& \cellcolor{blue!10}\textbf{71.82} & \cellcolor{blue!10}\textbf{60.12}
& \cellcolor{blue!10}\underline{55.18} & \cellcolor{blue!10}\textbf{93.50} \\
\cline{2-13}
& \multirow{3}{*}{40\%} & Wanda-sp & 32.36 & 24.71 & 16.75 & \underline{1092.31} &\underline{68.09} &23.60 &45.46 &32.21 &10.22 &55.23 \\
&  & FLAP &\textbf{45.95} &\underline{40.98} &\underline{19.33} &798.45 &52.29 &\underline{49.28} &\underline{54.50} &\underline{35.34} &\underline{47.72} &\underline{61.51} \\

& & \cellcolor{blue!10}SlimVLM
& \cellcolor{blue!10}\underline{45.76} & \cellcolor{blue!10}\textbf{47.14}
& \cellcolor{blue!10}\textbf{36.78} & \cellcolor{blue!10}\textbf{1245.70}
& \cellcolor{blue!10}\textbf{85.25} & \cellcolor{blue!10}\textbf{53.77}
& \cellcolor{blue!10}\textbf{60.54} & \cellcolor{blue!10}\textbf{42.88}
& \cellcolor{blue!10}\textbf{40.66} & \cellcolor{blue!10}\textbf{74.59} \\
\bottomrule
\end{tabular}
}
\label{llava-main-result}
\end{table*}

\begin{figure*}[t]
  \centering
  \includegraphics[width=\linewidth]{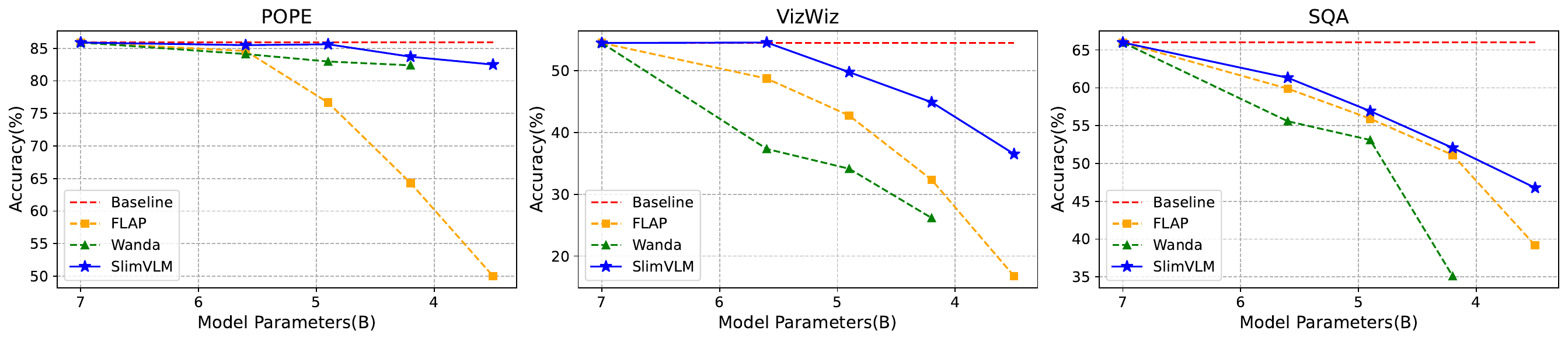}

   \caption{Performance of SlimVLM on the LLaVA-1.5 across three multimodal benchmarks. }
   \label{fig2}
\end{figure*}

\subsection{Main Results}

\textbf{Pruning Result on Various Tasks.} Table \ref{llava-main-result} presents a comparison of the experimental results between SlimVLM and the baseline methods on LLaVA-1.5-7B and LLaVA-Next-7B. All methods use the same visual token selection strategy. The experiments were conducted across 8 public multimodal benchmark datasets, reporting the performance for each task and providing the average accuracy compared to the unpruned models. When applying a 20\% pruning rate to the LLaVA-1.5 model, SlimVLM demonstrates the best performance advantage in the vast majority of tasks, with only suboptimal performance on the VQA-v2 task. In several benchmarks, SlimVLM demonstrates superior generalization capabilities and adaptability, including VizWiz(\textbf{54.55} vs 48.72 vs 37.34), MMB(\textbf{56.44} vs 52.58 vs 51.72). Furthermore, our SlimVLM achieves performance on par with the vanilla model on both the POPE and VizWiz benchmarks. 

\renewcommand{\arraystretch}{1.2} 
\setlength\heavyrulewidth{0.25ex}
\begin{table}[t!]
\centering
\small
\vspace{-5pt}
\caption{Performance of SlimVLM on Qwen2.5-VL-7B across Multiple Multimodal Benchmarks. In our experiments, we set the max pixels to $1280 \times 28 \times 28$.}
\resizebox{\textwidth}{!}{%
\begin{tabular}{c|ccccccccc|c}
\toprule
Ratio  & GQA  & MMB & MMB$^{\text{CN}}$ & MME & POPE & SQA$^{\text{I}}$ & VQA$^{\text{2}}$ & VQA$^{\text{T}}$ & VizWiz & Average(\%)  \\ \cline{1-11}    
 0\% &58.28 &80.67 &62.16 &2218.06 &87.82 &80.61 &76.42 &76.80 &66.86 &100  \\ 
 20\% &54.47 & 75.23 & 55.32 & 2033.76 & 86.78 & 73.34 & 72.48 & 69.55 & 64.27 & 93.19 \\ 
40\% &43.86 & 61.78 & 43.96 & 1726.55 & 84.37 & 64.32 & 63.79 & 57.18 & 42.68 & 77.56 \\
\bottomrule
\end{tabular}}
\vspace{-10pt}
\label{qwen-main-result}
\end{table}

\renewcommand{\arraystretch}{1.2} 
\setlength\heavyrulewidth{0.25ex}
\begin{table*}[tb!]
\centering
\small
\caption{Performance recovery of pruned LLaVA-1.5-7B via LoRA fine-tuning $(r=32)$}
\resizebox{\textwidth}{!}{%
\begin{tabular}{l|c|ccccccccc|c}
\toprule
Ratio & Tune & GQA  & MMB & MMB$^{\text{CN}}$ & MME & POPE & SQA$^{\text{I}}$ & VQA$^{\text{2}}$ & VQA$^{\text{T}}$ & VizWiz & Average(\%)  \\ \midrule   

0\% & - & \cellcolor{gray!20}61.94 & \cellcolor{gray!20}62.02
& \cellcolor{gray!20}50.26 & \cellcolor{gray!20}1780.64
& \cellcolor{gray!20}85.89 & \cellcolor{gray!20}65.99
& \cellcolor{gray!20}69.56 & \cellcolor{gray!20}48.07
& \cellcolor{gray!20}54.48 & \cellcolor{gray!20}100 \\ \midrule
\multirow{2}{*}{20\%} & $\times$ & 58.00&56.44&45.88&1682.96&85.49&61.33&65.90&41.56&54.55&93.80\\

& $\checkmark$ & \cellcolor{blue!10}59.67 & \cellcolor{blue!10}58.99
& \cellcolor{blue!10}48.26 & \cellcolor{blue!10}1712.84
& \cellcolor{blue!10}85.45 & \cellcolor{blue!10}65.18
& \cellcolor{blue!10}67.18 & \cellcolor{blue!10}44.38
& \cellcolor{blue!10}54.66 & \cellcolor{blue!10}96.80 \\ \midrule
\multirow{2}{*}{40\%} & $\times$ & 46.44& 43.57& 33.81& 1331.00& 83.71& 52.06& 54.10& 25.31& 44.89& 75.16 \\

& $\checkmark$ & \cellcolor{blue!10}51.28 & \cellcolor{blue!10}46.63
& \cellcolor{blue!10}42.78 & \cellcolor{blue!10}1511.74
& \cellcolor{blue!10}84.33 & \cellcolor{blue!10}58.64
& \cellcolor{blue!10}57.19 & \cellcolor{blue!10}33.17
& \cellcolor{blue!10}48.95 & \cellcolor{blue!10}84.01 \\
\bottomrule
\end{tabular}
}
\vspace{-8pt}
\label{lora_}
\end{table*}

Based on a comprehensive evaluation across all benchmarks, our SlimVLM maintains robust performance with an average score decrease of only 6.2\% compared to the original model. More importantly, it outperforms existing approaches by achieving an average score that is 4.43\% higher than FLAP and 8.49\% higher than Wanda-sp.

At a 40\% pruning rate, the advantage of our method becomes more pronounced. While the average performance decrease of our method is 24.84\%, it still maintains competitive performance on the vast majority of tasks. A notable exception is TextVQA, where its performance drops to 52.65\% of the vanilla model. In contrast, both FLAP and Wanda-sp suffer from severe performance collapse on certain datasets: FLAP achieves only 44.71\% of the vanilla model's performance on MME, while Wanda-sp drops to 38.10\% on MMB. Overall, the average scores of FLAP and Wanda-sp across all datasets are 62.15\% and 60.25\% respectively. These results demonstrate that our method's advantage becomes even more pronounced at higher pruning rates.

Figure \ref{fig2} visualizes the performance of LLaVA-1.5 across different pruning ratios on three tasks: POPE, VizWiz, and SQA. While FLAP and SlimVLM achieve comparable performance on the SQA task at pruning ratios up to 40\%, a noticeable performance gap emerges at the 50\% pruning level. Under this aggressive pruning ratio, Wanda-sp loses its ability to answer questions. These results demonstrate that our method maintains competitive performance even under aggressive pruning settings, exhibiting superior robustness compared to existing approaches.

\textbf{Pruning in Higher Resolution Models.} We also present the pruning results for LLaVA-Next-7B in Table \ref{llava-main-result}. Compared to LLaVA-1.5, LLaVA-Next processes higher-resolution images, producing up to 2880 visual tokens. We evaluated SlimVLM against other methods under two pruning ratios. At 20\% pruning, our method achieved an average score drop of only 6.5\%, outperforming FLAP and Wanda-sp by 3.55\% and 9.04\%, respectively. Even at a 40\% pruning ratio, our approach maintained strong performance with an average score of 74.59\%, significantly surpassing all baselines. Under the same setting, Wanda-sp suffered from severe performance degradation on several datasets; for example, on VizWiz, it attained only 17.38\% of the upper-bound accuracy, whereas our method reached 69.17\%. These results confirm the effectiveness of our approach in high-resolution visual settings.

\textbf{Pruning results on different architecture models.} Our method remains effective for models employing Grouped-Query Attention. Unlike traditional Multi-Head Attention, Grouped-Query Attention allows multiple query heads to share a common set of key-value heads. When pruning such models, we preserve heterogeneity in the distribution of query heads within each group after pruning, thereby better adapting to the structural characteristics of Grouped-Query Attention. As shown in Table \ref{qwen-main-result}, experimental results on the Qwen2.5-VL model demonstrate that our method maintains an average score of 93.19 at a 20\% pruning ratio and still achieves 77.56\% even at a 40\% pruning ratio. These results confirm that our approach can be effectively applied to diverse model architectures.

\renewcommand{\arraystretch}{1.2} 
\setlength\heavyrulewidth{0.25ex}
\begin{table*}[tb!]
\centering
\small
\caption{Ablation study on the token selection strategy, where "None" refers to using all visual tokens, "Random" refers to randomly selecting a number of visual tokens equal to the number of text tokens, "Fixed" refers to retain the top-k most important visual tokens, with k set to match the number of text tokens. }
\resizebox{\textwidth}{!}{%
\begin{tabular}{c|c|ccccccccc|c}
\toprule
Method & Select Strategy & GQA  & MMB & MMB$^{\text{CN}}$ & MME & POPE & SQA$^{\text{I}}$ & VQA$^{\text{2}}$ & VQA$^{\text{T}}$ & VizWiz & Average(\%)  \\\midrule   
\multirow{3}{*}{FLAP} & \cellcolor{gray!20}None
& \cellcolor{gray!20}46.64 & \cellcolor{gray!20}46.99
& \cellcolor{gray!20}28.44 & \cellcolor{gray!20}1174.77
& \cellcolor{gray!20}76.68 & \cellcolor{gray!20}48.69
& \cellcolor{gray!20}53.90 & \cellcolor{gray!20}36.11
& \cellcolor{gray!20}52.88 & \cellcolor{gray!20}76.26 \\ 
&Random & 52.94 & 50.10& 37.68 &1286.12&82.46&57.30&52.75&38.48&46.77&82.00 \\

& \cellcolor{blue!10}Ours
& \cellcolor{blue!10}55.45 & \cellcolor{blue!10}52.58
& \cellcolor{blue!10}41.24 & \cellcolor{blue!10}1645.92
& \cellcolor{blue!10}84.59 & \cellcolor{blue!10}59.89
& \cellcolor{blue!10}64.32 & \cellcolor{blue!10}40.59
& \cellcolor{blue!10}48.32 & \cellcolor{blue!10}89.37 \\ \midrule
\multirow{4}{*}{SlimVLM} & \cellcolor{gray!20}None
& \cellcolor{gray!20}47.09 & \cellcolor{gray!20}35.57
& \cellcolor{gray!20}22.08 & \cellcolor{gray!20}1065.51
& \cellcolor{gray!20}83.97 & \cellcolor{gray!20}52.70
& \cellcolor{gray!20}57.06 & \cellcolor{gray!20}37.01
& \cellcolor{gray!20}46.80 & \cellcolor{gray!20}73.30 \\ 
&Random & 58.17 &55.50 & 42.61 & 1488.59 & 84.67 &59.25 & 65.76 & 41.15 & 46.45 & 89.51 \\
&Fixed & 57.52 & 56.01 & 43.99 & 1683.80 & 85.28 & 60.93 & 65.70 & 41.01 & 47.92 & 91.63 \\
 
& \cellcolor{blue!10}Ours
& \cellcolor{blue!10}58.00 & \cellcolor{blue!10}56.44
& \cellcolor{blue!10}45.88 & \cellcolor{blue!10}1682.96
& \cellcolor{blue!10}85.49 & \cellcolor{blue!10}61.33
& \cellcolor{blue!10}65.90 & \cellcolor{blue!10}41.56
& \cellcolor{blue!10}54.55 & \cellcolor{blue!10}93.80 \\
\bottomrule
\end{tabular}
}
\vspace{-10pt}
\label{ablition1}
\end{table*}

\renewcommand{\arraystretch}{1.2} 
\setlength\heavyrulewidth{0.25ex}
\begin{table*}[tb!]
\centering
\small
\caption{Ablation study on the dynamic pruning rates for modules.}
\resizebox{\textwidth}{!}{%
\begin{tabular}{c|ccccccccc|c}
\toprule
Method  & GQA  & MMB & MMB$^{\text{CN}}$ & MME & POPE & SQA$^{\text{I}}$ & VQA$^{\text{2}}$ & VQA$^{\text{T}}$ & VizWiz & Average(\%)  \\ \midrule  
w/o dynamic pruning rates & 56.77&54.73&42.78&1677.93&85.66&59.99&66.60&40.10&44.89&90.16 \\ 

 w/ dynamic pruning rates &58.00 & 56.44 & 45.88 & 1682.96 &  85.49 & 61.33 & 65.90 &  41.56 & 54.55  & 93.80 \\
\bottomrule
\end{tabular}
}
\vspace{-10pt}
\label{ablition2}
\end{table*}

\begin{table*}[t]
\centering
\begin{minipage}{0.5\linewidth}
\renewcommand{\arraystretch}{1.2}
\setlength{\tabcolsep}{2pt}
\centering
\caption{Model performance and memory usage under different development set sampling sizes on GQA, MMBench, MME, and SQA.}
\label{sample_main}
\resizebox{\linewidth}{!}{
\begin{tabular}{c|cccc|c}
\toprule
Samples & GQA & MMB & MME & SQA$^{\text{I}}$ & Memory(G) \\ \hline
12 & 57.88 & 55.84& 1695.25 &61.68 & 17.74\\
36 & 56.63& 57.13 & 1578.76 & 59.59 & 26.11\\ \hline
24 & 58.00 & 56.44 & 1682.92 & 61.33 & 21.14 \\
\bottomrule
\end{tabular}
}
\end{minipage}
\hfill
\begin{minipage}{0.48\linewidth}
\renewcommand{\arraystretch}{1.15}
\setlength{\tabcolsep}{3pt}
\centering
\caption{Evaluation results of the inference speed before and after pruning.}
\label{inference}
\resizebox{\linewidth}{!}{
\begin{tabular}{c|c|c|c}
\toprule
Ratio & Method & Tokens/s & Speed up \\ \hline
0\% & Vanilla & 85.39 & 1$\times$ \\ \hline
\multirow{3}{*}{20\%} 
& Wanda-sp & 94.76 ($\uparrow$ 10.97\%) & 1.11$\times$ \\
& FLAP & 93.33 ($\uparrow$ 9.30\%) & 1.09$\times$ \\
& SlimVLM & 106.69 ($\uparrow$ 24.94\%) & 1.25$\times$ \\
\bottomrule
\end{tabular}
}
\end{minipage}
\end{table*}

\textbf{Performance Recovery After Pruning}
Considering the performance degradation caused by the reduction in module quantity after pruning, especially under high pruning ratios, we performed performance recovery training on the pruned model. Specifically, we employed the LoRA fine-tuning strategy and retrained the pruned model using the MobileVLM~\cite{mobilevlm} dataset to mitigate the performance loss induced by pruning. As illustrated in Table \ref{lora_}, the post-training recovery yields a 3\% performance restoration at a 20\% pruning ratio. Notably, under a higher pruning ratio of 40\%, the recovery effect becomes more pronounced, achieving a substantial 8.85\% performance improvement.

\subsection{Ablation Studies}

To verify the effectiveness of our proposed method, we conducted ablation studies on LLaVA-1.5-7B to validate each of our proposed strategies. 
Unless otherwise specified, all experiments were conducted at a pruning ratio of 20\%.

\begin{wrapfigure}[16]{r}{0.48\columnwidth}
  \vspace{-22pt}
  \centering
  \includegraphics[width=\linewidth]{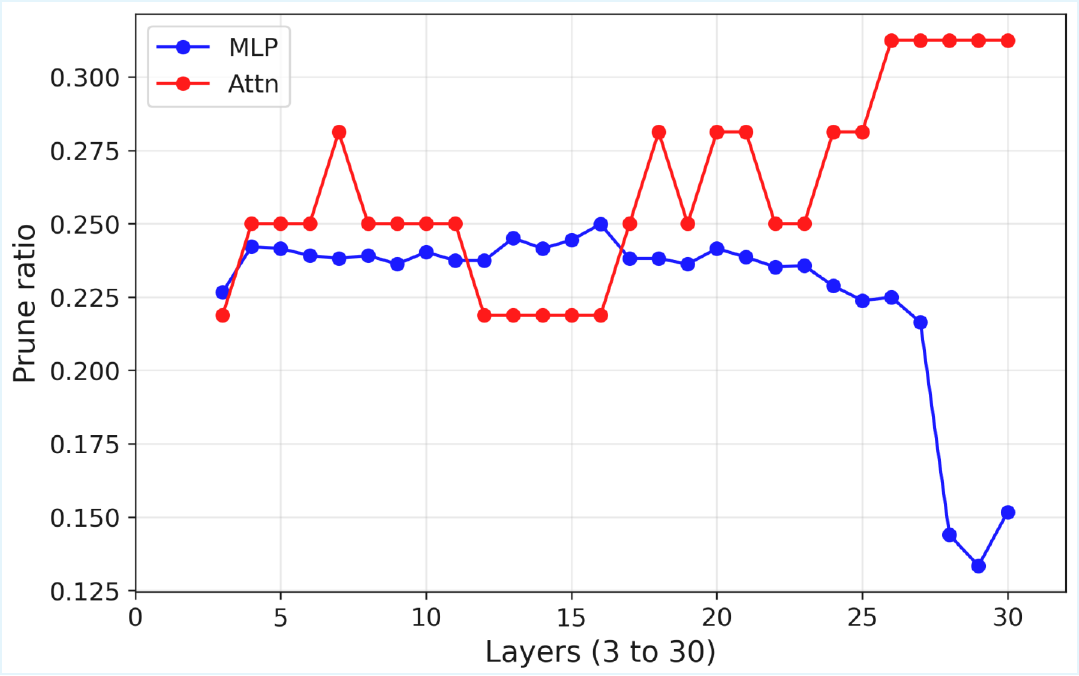}
  \caption{The distribution of dynamic pruning rates in LLaVA-1.5: Attention Heads and MLP Channels under 20\% pruning ratio.}
  \label{prune_ratio}
\end{wrapfigure}
\noindent
\textbf{Adaptive visual token selection strategy.} From Table \ref{ablition1}, we assessed the efficacy of the adaptive visual token selection strategy by comparing FLAP and our proposed SlimVLM under three different settings: using all visual tokens (None), randomly selecting a number of visual tokens equal to the text tokens (Random), and employing our proposed approach (Ours). When integrated with the adaptive token selection strategy, both pruning methods achieved significantly higher performance than using random or no token selection. This substantial gain is attributed to the fact that both methods rely on module importance calculations that are sensitive to visual tokens. FLAP identifies modules to prune based on the output fluctuation before and after pruning, while SlimVLM determines redundant modules using the Pearson correlation coefficient. Although these pruning metrics have been proven effective, their judgment during the pruning process can be compromised when the input contains a large number of redundant tokens. Furthermore, we compared our method against a top-k selection strategy(Fixed), where k is set to the number of text tokens. The results demonstrate that our proposed adaptive token selection strategy leads to superior performance.

\textbf{Dynamic Pruning Rates for Modules.} We compute the Pearson correlation coefficient between the pre- and post-pruning outputs of MHA and MLP layers, and then dynamically adjust the pruning ratio according to Equation \ref{eq:9}. As shown in Table \ref{ablition2}, which compares SlimVLM with and without dynamic pruning, the results clearly demonstrate the effectiveness of our dynamic pruning strategy. 
As shown in Figure \ref{prune_ratio}, we visualized the pruning ratios of different modules at each layer under a 20\% pruning ratio. We observe that deeper MLP layers possess lower sparsity tolerance, necessitating the retention of more channels to maintain performance. Meanwhile, attention heads in intermediate layers similarly exhibit limited sparsity tolerance and require more heads to be preserved, while deeper layers enable more aggressive head pruning.

\textbf{Impact of Hyperparameters in Equation \ref{eq:9}}
In Equation~\ref{eq:9}, we introduce two hyperparameters, $\gamma$ and $T$, to control the sensitivity of dynamic pruning ratio adjustment. We conduct additional experiments to evaluate the impact of different hyperparameter settings, as shown in Table~\ref{tab:hyperparameter}. The results show that the proposed method remains stable across different choices of $\gamma$ and $T$, with average performance varying by less than 3\%. This suggests that SlimVLM is robust to these hyperparameters and does not require extensive retuning.

\renewcommand{\arraystretch}{1.2} 
\setlength\heavyrulewidth{0.25ex}
\begin{table*}[tb!]
\centering
\small
\caption{Impact of the hyperparameters $\gamma$ and $T$ in Equation~\ref{eq:9}. In our main experiments, $\gamma$ and $T$ are set to 5 and 1, respectively, and this configuration is used as the reference baseline with the average performance normalized to 100\%.}
\resizebox{\textwidth}{!}{%
\begin{tabular}{c|c|ccccccccc|c}
\toprule
$\gamma$ & T & GQA  & MMB & MMB$^{\text{CN}}$ & MME & POPE & SQA$^{\text{I}}$ & VQA$^{\text{2}}$ & VQA$^{\text{T}}$ & VizWiz & Average(\%)  \\\midrule   
1 & 1 & 56.32 & 55.21 & 44.24 & 1619.79 & 85.46 & 57.24 & 64.46 & 41.29 & 53.45 & 97.25 \\
3 & 1 & 56.40 & 55.84 & 44.59 & 1623.69 & 85.43 & 58.29 & 65.15 & 41.48 & 53.57 & 97.93 \\
\rowcolor{blue!10} 5 & 1 & 58.00 & 56.44 & 45.88 & 1682.69 & 85.49 & 61.33 & 65.90 & 41.56 & 54.55 & 100 \\
5 & 0.95 & 57.38 & 56.07 & 45.18 & 1646.96 & 85.43 & 59.79 & 64.62 & 41.23 & 54.07 & 98.66 \\
5 & 0.9 & 57.19 & 55.87 & 44.73 & 1595.14 & 85.44 & 59.38 & 64.98 & 40.99 & 53.99 & 97.98 \\
\bottomrule
\end{tabular}
}
\vspace{-10pt}
\label{tab:hyperparameter}
\end{table*}

\textbf{Impact of Development Set Size}
In our experiments, module importance was computed on a development set consisting of 24 randomly sampled instances from each of the English and Chinese versions of MMBench. Additionally, we conducted experiments with alternative sample sizes of 12 and 36. Table \ref{sample_main} reports the model’s performance and memory usage across the GQA, MMBench, MME, and SQA benchmarks under different sampling sizes.

\subsection{Inference Speed}
This section empirically compares the inference efficiency of different pruning methods.  Table \ref{inference} demonstrates that our proposed method achieves remarkable inference acceleration, with a 24.94\% speed improvement at a 20\% pruning rate. In contrast, although FLAP and Wanda-sp also adopt structured pruning strategies, their actual acceleration performance is relatively weaker due to suboptimal allocation of pruning ratios across attention heads and MLP layers.

\section{Conclusion}
In this paper, we introduce SlimVLM, a structured pruning method specifically designed for VLMs. To address interference from redundant visual tokens in module importance estimation, we propose an adaptive token selection strategy that filters them out before importance calculation. Specifically, we reuse the decoder layer's attention map and leverage text queries to compute visual token importance scores, discarding those below a predefined threshold. Furthermore, we introduce a sensitivity-aware dynamic pruning strategy based on Pearson correlation coefficients for both attention heads and channels, which quantifies linear output discrepancies to dynamically allocate pruning ratios across modules. Extensive experiments demonstrate that SlimVLM achieves strong performance across various architectures and pruning ratios.

\nocite{langley00}

\bibliographystyle{unsrtnat}
\bibliography{example_paper}

\newpage

\section{Appendix}
\appendix

\begin{figure}[!htbp]
  \centering
  \begin{minipage}[t]{0.42\linewidth}
    \centering
    \includegraphics[width=\linewidth]{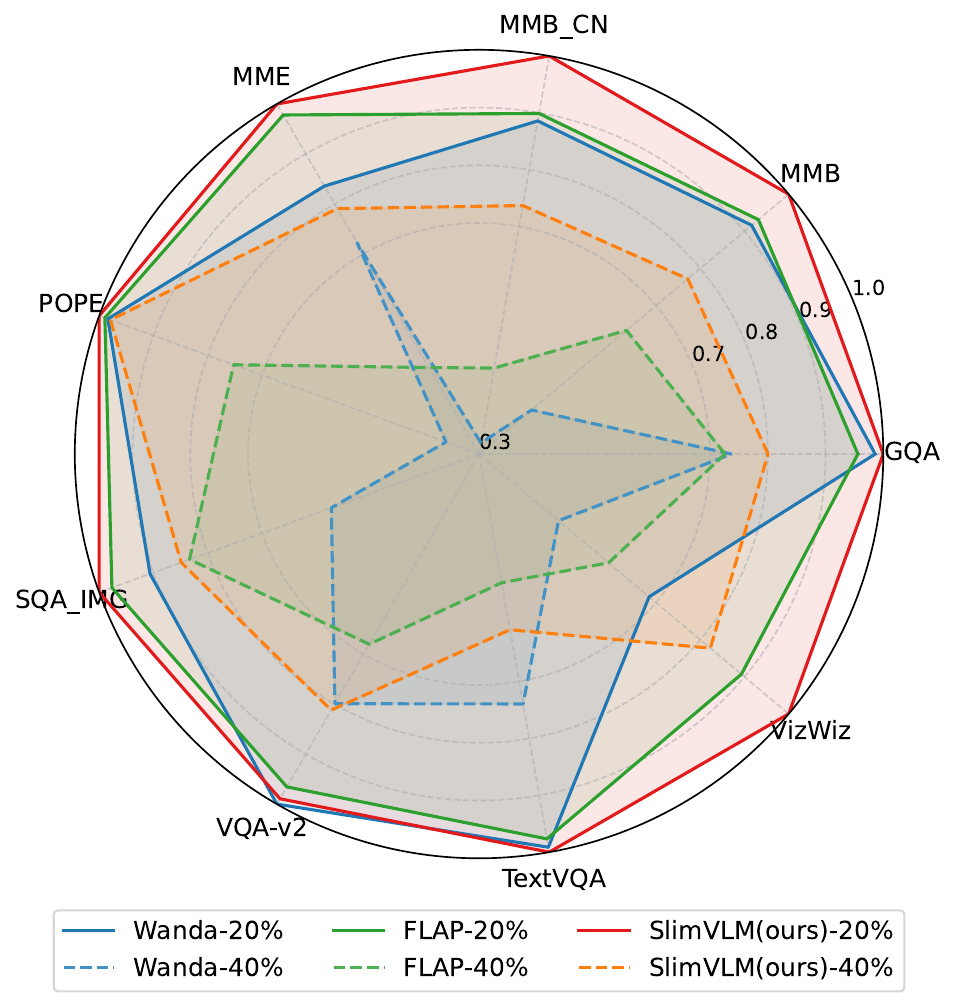}
    \caption{Comparative analysis of model performance.}
    \label{radar_chart}
  \end{minipage}
  \hfill  
  \begin{minipage}[t]{0.48\linewidth}
    \centering
    \includegraphics[width=\linewidth]{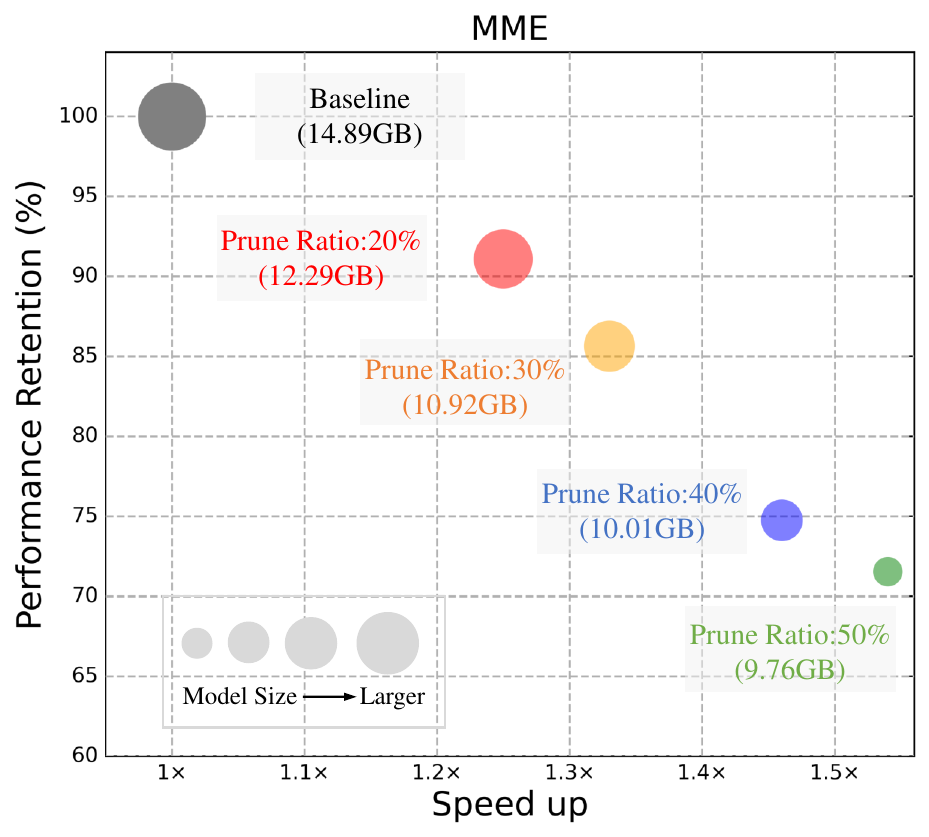}
    \caption{Accuracy, inference speed, and GPU memory usage of SlimVLM on MME benchmark across different pruning ratios.}
    \label{MMEGPU}
  \end{minipage}
\end{figure}

\section{Benchmark Datasets}

We conducted experiments on several widely used visual understanding benchmarks. The specific metrics and system prompt for these benchmarks are presented in Table \ref{benchmark_detail}.

\textbf{GQA~\cite{GQA}.} The GQA dataset comprises three key components: scene graphs, questions, and images. The image section includes not only the images themselves but also their spatial features along with features of all the objects present within them. The questions in GQA are specifically designed to evaluate a model's comprehension of visual scenes and its capacity for reasoning about various aspects of an image.

\textbf{MMBench~\cite{MMBench}.} The MMBench benchmark is designed to comprehensively evaluate a model's overall capabilities through a multi-dimensional hierarchical framework. This framework comprises three distinct levels: the first level (L-1) outlines two core abilities—Perception and Reasoning. Building upon this, the second level (L-2) expands into six sub-abilities. These are further refined into twenty specific ability dimensions at the third level (L-3). Such a granular structure allows for a thorough and detailed assessment of the model's diverse competencies.

\textbf{MME~\cite{MME}.} The MME benchmark is a comprehensive evaluation framework meticulously designed to assess a model's performance across diverse dimensions. It encompasses 14 subtasks tailored to evaluate both perceptual and cognitive abilities. By leveraging manually curated instruction-answer pairs and a concise instruction design, MME effectively mitigates issues such as data leakage and ensures a fairer evaluation of model capabilities.

\textbf{POPE~\cite{POPE}.} The POPE benchmark is designed to quantify object hallucination in vision-language models. It formulates the evaluation as a series of binary questions about object presence in images, requiring models to answer them. The benchmark employs Accuracy, Recall, Precision, and F1 score as its metrics, which are computed under three distinct sampling strategies to provide a precise assessment of hallucination tendencies.

\textbf{ScienceQA~\cite{SQA}.} The ScienceQA benchmark encompasses a wide array of domains, spanning natural science, language science, and social science. Questions are organized hierarchically: first by topic, then by category, and finally by skill. This structure yields 26 topics, 127 categories, and 379 distinct skills, offering a rich and diverse set of scientific questions. ScienceQA thereby provides a comprehensive evaluation of model capabilities in multimodal understanding, multi-step reasoning, and interpretability.

\textbf{VQA-v2~\cite{VQAv2}.} The VQA-v2 benchmark assesses models' visual perception capabilities using open-ended questions. It comprises 265,016 images depicting diverse real-world scenes and objects, thereby providing rich visual contexts. For each question, ten ground-truth answers are collected from human annotators, enabling a robust and comprehensive evaluation of model accuracy.

\textbf{TextVQA~\cite{TextVQA}.} The TextVQA benchmark is designed to evaluate a model's ability to integrate and reason about diverse textual information within images. Through a series of visual question-answering tasks that incorporate rich textual cues, it assesses both text understanding and reasoning capabilities. To perform accurately, models must not only interpret visual content but also read and reason over the text present in the images.

\textbf{VizWiz~\cite{VizWiz}.} The VizWiz benchmark is derived from a natural visual question-answering scenario involving individuals who are blind. It comprises over 31,000 visual questions, each originating from a blind person who captured an image using a mobile phone and recorded a corresponding spoken question. Each question is accompanied by 10 crowdsourced answers. This benchmark was established to encourage the broader community to develop generalized algorithms capable of assisting blind individuals.

\renewcommand{\arraystretch}{1.2} 
\setlength\heavyrulewidth{0.25ex}
\begin{table*}[htbp!]
\centering
\small

\caption{Overview of the Benchmarks Employed in this Study}

\resizebox{\textwidth}{!}{%
\begin{tabular}{c|c|c}
\toprule
Benchmark & Metric & System Prompt \\ \hline
GQA~\cite{GQA} & exact match & Answer the question using a single word or phrase. \\
MMBench~\cite{MMBench} & gpt eval score & Answer with the option's letter from the given choices directly. \\
MME~\cite{MME} & perception score,cognition score & Answer the question using a single word or phrase. \\
POPE~\cite{POPE} & accuracy & Answer the question using a single word or phrase.\\
ScienceQA~\cite{SQA} & exact match & Answer with the option's letter from the given choices directly.\\
VQA-v2~\cite{VQAv2} & exact match & Answer the question using a single word or phrase.\\
TextVQA~\cite{TextVQA} & exact match & Answer the question using a single word or phrase.\\
VizWiz~\cite{VizWiz} & exact match & Answer the question using a single word or phrase.\\
\bottomrule
\end{tabular}
}

\label{benchmark_detail}
\end{table*}

\section{Extended Experiment Results}
\subsection{Performance Comparison}
As shown in Figure \ref{radar_chart}, we compare SlimVLM against two baseline models, wanda-sp and FLAP, across multiple datasets. SlimVLM demonstrates superior performance, an advantage that becomes even more pronounced at higher pruning ratios (40\%).

\begin{figure*}[htbp!]
  \centering
  \includegraphics[width=\linewidth]{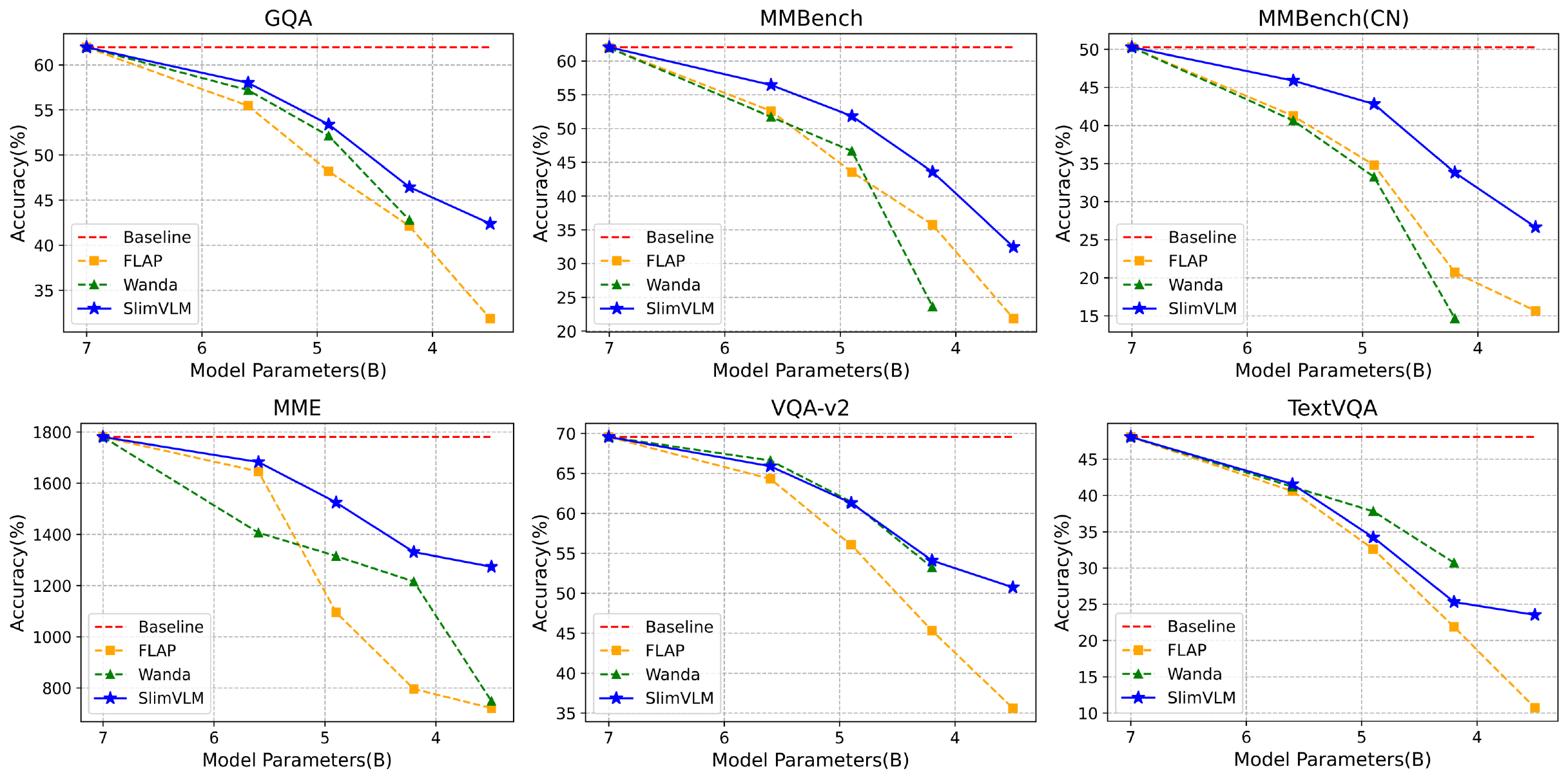}

   \caption{Performance of SlimVLM on the LLaVA-1.5 across six multimodal benchmarks. The horizontal axis denotes the number of model parameters, and the vertical axis presents the evaluation accuracy for each dataset.}
   \label{suppl_performence}
\end{figure*}

Figure \ref{suppl_performence} visualizes the performance of LLaVA-1.5 across different pruning ratios on six tasks: GQA, MMBench, MMBench(CN), MME, VQA-v2 and TextVQA. These visualizations provide deeper insights into the performance of SlimVLM across different benchmarks and pruning ratios, in comparison with two baselines: Wanda-sp and FLAP.

\subsection{Comparison with Additional Pruning Baselines}
To further strengthen the comparison with structured pruning methods, we include additional LLM pruning baselines in Table \ref{tab:additional_structured_pruning}, including LLM-Pruner, LoRAP, Puzzle, and SlimLLM. Under the same 20\% pruning ratio on LLaVA-1.5-7B, SlimVLM consistently achieves the best average performance, outperforming the strongest additional baseline SlimLLM by 3.28\% in average relative performance. This suggests that explicitly accounting for visual-token interference during pruning is important for adapting LLM structured pruning methods to VLMs.

\renewcommand{\arraystretch}{1.2}
\setlength\heavyrulewidth{0.25ex}
\begin{table*}[tb!]
\centering
\small
\caption{\textbf{Performance comparison with additional structured pruning baselines on LLaVA-1.5-7B at a 20\% pruning ratio.} The “Average” metric is computed as the mean relative performance across datasets, normalized by the corresponding vanilla model performance. The \textbf{"Bolded"} represents the best result under the same pruning ratio, and "\underline{underline}" denotes the second best.}
\resizebox{\textwidth}{!}{%
\begin{tabular}{l|ccccccccc|c}
\toprule
Method & GQA & MMB & MMB$^{\text{CN}}$ & MME & POPE & SQA$^{\text{I}}$ & VQA$^{\text{2}}$ & VQA$^{\text{T}}$ & VizWiz & Average(\%) \\
\midrule
\rowcolor{gray!20} Vanilla & 61.94 & 62.02 & 50.26 & 1780.64 & 85.89 & 65.99 & 69.56 & 48.07 & 54.48 & 100.00 \\
Wanda-sp & 57.19 & 51.72 & 40.64 & 1406.53 & 84.12 & 55.58 & \textbf{66.62} & \underline{41.20} & 37.34 & 85.31 \\
FLAP & 55.45 & 52.58 & 41.24 & \underline{1645.92} & 84.59 & \underline{59.89} & 64.32 & 40.59 & 48.72 & 89.37 \\
LLM-Pruner & 56.14 & 50.37 & 41.11 & 1562.36 & 84.33 & 57.67 & 63.47 & 41.08 & 42.57 & 86.87 \\
LoRAP & 56.71 & 52.34 & 41.89 & 1588.37 & 84.70 & 58.15 & 63.18 & 40.17 & 44.36 & 87.89 \\
Puzzle & \underline{57.34} & \underline{55.03} & \underline{43.17} & 1569.41 & 84.44 & 59.30 & 63.54 & 39.47 & 46.61 & 89.17 \\
SlimLLM & 57.21 & 54.78 & 42.01 & 1611.38 & \underline{84.72} & 59.17 & 64.57 & 40.88 & \underline{51.07} & \underline{90.52} \\
\rowcolor{blue!10} \textbf{SlimVLM} & \textbf{58.00} & \textbf{56.44} & \textbf{45.88} & \textbf{1682.96} & \textbf{85.49} & \textbf{61.33} & \underline{65.90} & \textbf{41.56} & \textbf{54.55} & \textbf{93.80} \\
\bottomrule
\end{tabular}%
}
\label{tab:additional_structured_pruning}
\end{table*}

\subsection{Scaling to Larger Models}
We further apply SlimVLM to LLaVA-1.5-13B, which is a larger vision-language model. These experiments follow the same setup described in Section "Experiment," and the results are presented in Table \ref{13b-result}.

We observe that the benefits of pruning become more pronounced on larger models. At a 20\% pruning ratio, our SlimVLM outperforms existing methods across the board, achieving the best performance on all evaluated tasks. In several benchmarks, SlimVLM demonstrates superior generalization capabilities and adaptability, including VizWiz(\textbf{59.86} vs 58.41 vs 57.77), MME(\textbf{1625.57} vs 1539.38 vs 1469.49). 

Based on a comprehensive evaluation across all benchmarks, our SlimVLM maintains robust performance, with an average score degradation of only 3.58\% compared to the original model—significantly more stable than the pruning results on the 7B-scale model. Specifically, SlimVLM obtains an average score that is 2.53\% higher than FLAP and 3.62\% higher than Wanda-sp. 

When the pruning ratio increases to 40\%, SlimVLM still maintains an average accuracy of 84.67\% relative to the original model, whereas Wanda-sp drops sharply to only 67.60\%. This suggests that Wanda-sp may suffer from severe performance degradation—or even catastrophic failure—under extreme pruning conditions.

\renewcommand{\arraystretch}{1.3} 
\setlength\heavyrulewidth{0.25ex}
\begin{table*}[htbp!]
\centering

\caption{\textbf{Performance of SlimVLM on LLaVA-1.5-13B across Multiple Multimodal Benchmarks}. The “Average” metric is computed as the mean relative performance across datasets, normalized by the corresponding vanilla model performance. The \textbf{"Bolded"} represents the best result under the same pruning ratio, and "\underline{underline}" denotes the second best.}

\resizebox{\textwidth}{!}{%
\begin{tabular}{c|c|c|ccccccccc|c}
\toprule
 & Ratio & Method & GQA  & MMB & MMB$^{\text{CN}}$ & MME & POPE & SQA$^{\text{I}}$ & VQA$^{\text{2}}$ & VQA$^{\text{T}}$ & VizWiz & Average(\%)  \\ 
 \midrule \multirow{8}{*}{\rotatebox[origin=c]{90}{\textcolor{black}{\textbf{LLaVA 1.5-13B}}}} %

&  0\% &  Vanilla & 61.85 & 66.07 & 54.81 & 1664.98 & 85.53 & 71.44 & 74.36 & 52.92 & 59.20 & 100 \\ \cline{2-13}

& \multirow{3}{*}{20\%}  & Wanda-sp & 58.06 & 61.17 & \underline{50.85} & 1469.49 & 78.97 & \underline{66.93} & \underline{69.02} & \underline{48.32} & 57.77  & 92.80   \\

&  & FLAP & \underline{59.04} & \textbf{63.14} & 50.43 & \underline{1539.38} & \underline{82.97} & 66.14 & 68.64 & 47.08 & \underline{58.41} & \underline{93.89}  \\

&  &  SlimVLM & \textbf{60.23} & \textbf{63.14} & \textbf{51.91} & \textbf{1625.57} & \textbf{84.32} & \textbf{67.72} & \textbf{70.62} & \textbf{49.24} & \textbf{59.86} & \textbf{96.42} \\

\cline{2-13}
& \multirow{3}{*}{40\%} & Wanda-sp & 42.92 & 41.32 & 24.07 & 1200.56 & \underline{78.07} & 46.60 & 51.32 & 33.23 & 42.67& 67.60  \\

&  & FLAP & \underline{50.02} & \underline{50.46} & \underline{33.42} & \underline{1375.61} & 76.51 & \underline{60.19} & \underline{60.43} & \underline{38.55} & \underline{55.30} & \underline{80.23}  \\

&  &  SlimVLM & \textbf{54.65} & \textbf{53.87} & \textbf{34.89} & \textbf{1442.90} & \textbf{83.72} & \textbf{63.21} & \textbf{62.86} & \textbf{38.93} & \textbf{57.61} & \textbf{84.67}      \\
\bottomrule
\end{tabular}
}

\label{13b-result}
\end{table*}

\subsection{Effect of Attention Score Ranking on Visual Token Selection}
Most existing studies show that redundant visual tokens in vision-language models (VLMs) typically receive lower attention scores during inference and have only a marginal impact on the overall model performance. The presence of a large number of such redundant tokens can further interfere with module importance estimation, potentially causing critical components to be mistakenly removed or redundant components to be unnecessarily retained. To further investigate how visual token importance affects structured pruning, we fix the token retention ratio and select visual tokens from the bottom 20\%, middle 20\%, and top 20\% of the attention score ranking for module importance estimation, respectively. We then evaluate the pruning performance under different selection strategies, as shown in Table~\ref{tab:attention-ranking}. 

\renewcommand{\arraystretch}{1.0}
\setlength\heavyrulewidth{0.25ex}
\begin{table*}[htbp!]
\centering
\small
\caption{Performance comparison when retaining visual tokens from different attention score ranking ranges for module importance estimation. Bottom 20\%, Middle 20\%, and Top 20\% denote selecting visual tokens from the corresponding attention-score ranges.}
\resizebox{\textwidth}{!}{%
\begin{tabular}{c|ccccccccc|c}
\toprule
Model  & GQA  & MMB & MMB$^{\text{CN}}$ & MME & POPE & SQA$^{\text{I}}$ & VQA$^{\text{2}}$ & VQA$^{\text{T}}$ & VizWiz & Average(\%)  \\ \midrule
Vanilla & 61.94 & 62.02 & 50.26 & 1780.64 & 85.89 & 65.99 & 69.56 & 48.07 & 54.48 & 100.00 \\
Bottom 20\% & 54.79 & 49.91 & 40.77 & 1530.37 & 84.05 & 55.38 & 61.25 & 38.17 & 42.58 & 84.82 \\
Middle 20\% & 56.58 & 53.69 & 42.16 & 1596.78 & 84.37 & 58.12 & 64.06 & 39.49 & 50.16 & 89.34 \\
\rowcolor{blue!10}
Top 20\% & 57.37 & 55.31 & 44.08 & 1658.73 & 85.53 & 60.71 & 65.18 & 40.90 & 53.82 & 92.42 \\
\bottomrule
\end{tabular}
}
\label{tab:attention-ranking}
\end{table*}

The results show that retaining the top 20\% visual tokens achieves the best post-pruning performance, improving the average performance by 7.6\% compared with retaining the bottom 20\% visual tokens. This demonstrates that attention scores are effective indicators of visual token utility, and selecting high-attention visual tokens helps obtain more accurate module importance estimates for structured pruning.

\subsection{Threshold for Token Selection.}
In our Adaptive Visual Token Selection module, visual tokens whose average attention score $p$ falls below a threshold $\alpha$ are filtered out. The threshold is defined as $\alpha = \text{mean}(p) \times \beta$. To analyze the sensitivity of this hyperparameter, we report the pruning results with $\beta$ values of $[0.5, 0.3, 0.2, 0.1, 0.05]$ in Table \ref{ablition3}. A higher $\beta$ value leads to a more aggressive pruning strategy by retaining fewer visual tokens for the subsequent computation of module importance.Experimental results show that the pruned model performs optimally when $\beta = 0.2$. Increasing or decreasing the $\beta$ value leads to a certain degree of performance degradation.

\renewcommand{\arraystretch}{1.0} 
\setlength\heavyrulewidth{0.25ex}
\begin{table*}[htbp!]
\centering
\small
\caption{Ablation Studies for the value of $\beta$.}
\resizebox{\textwidth}{!}{%
\begin{tabular}{c|ccccccccc|c}
\toprule
$\beta$  & GQA  & MMB & MMB$^{\text{CN}}$ & MME & POPE & SQA$^{\text{I}}$ & VQA$^{\text{2}}$ & VQA$^{\text{T}}$ & VizWiz & Average(\%)  \\ \cline{1-11}    
0.05 & 46.00&52.34&38.62&1060.14&82.41&55.38&56.02&36.90&50.97&80.64 \\
0.1 & 53.41&51.72&36.68&1474.75&85.84&60.78&61.72&38.21&46.93 &85.76\\
 0.2 &58.00&56.44&45.88&1682.96&85.49&61.33&65.90&41.56&54.55&93.80 \\
0.3 &57.93&54.98&42.44&1526.02&85.23&59.94&66.12&39.98&48.39&89.94 \\
0.5&57.94&54.81&41.24&1483.12&84.69&59.84&64.26&38.10&41.14&87.08 \\
\bottomrule
\end{tabular}
}
\label{ablition3}
\end{table*}

\subsection{Comparison with VLM Token Compression Methods}

Current VLM pruning mainly focuses on token pruning, which differs from the approach of Our SlimVLM. While token pruning primarily emphasizes reducing the number of input tokens or adjusting token representations to enhance computational efficiency, SlimVLM is designed to reduce the model's deployment parameter size, a crucial factor for edge deployment. We compared SlimVLM with two state-of-the-art token pruning methods: FastV and PDrop. As shown in Table \ref{token_compress}, the model pruned with SlimVLM achieves performance on par with FastV. Although its performance is slightly lower than that of PDrop, SlimVLM still manages to achieve the best performance after post-training.

\renewcommand{\arraystretch}{1.2} 
\setlength\heavyrulewidth{0.25ex}
\begin{table*}[htbp!]
\centering
\small
\caption{Performance comparison with visual token compression baselines on LLaVA-1.5-7B. $\dag$ indicates LoRA-based recovery.}
\resizebox{\textwidth}{!}{%
\begin{tabular}{l|ccccccccc|c}
\toprule
Method  & GQA  & MMB & MMB$^{\text{CN}}$ & MME & POPE & SQA$^{\text{I}}$ & VQA$^{\text{2}}$ & VQA$^{\text{T}}$ & VizWiz & Average(\%)  \\ \cline{1-11}    
\rowcolor{gray!20}Vanilla & 61.94 & 62.02 & 50.26 & 1780.64 & 85.89 & 65.99 & 69.56 & 48.07 & 54.48 & 100 \\
FastV & 52.70 & 58.18 & 49.05 & 1560.68 & 80.70 & 63.88 & 65.78 & 43.12 & 55.54 & 93.46\\
PDrop & 57.38 & 57.14 & 47.17 & 1691.47 & 82.30 & 64.84 & 65.77 & 45.66 & 54.33 & 95.22 \\
 SlimVLM & 58.00 & 56.44 & 45.88 & 1682.96 & 85.49 & 61.33 & 65.90 & 41.56 & 54.55 & 93.80\\
\rowcolor{blue!10}$\text{SlimVLM}^\dag$ & 59.67 & 58.99 & 48.26 & 1712.84 & 85.45 & 65.18 & 67.18 & 44.38 & 54.66 & 96.80 \\
\bottomrule
\end{tabular}
}
\label{token_compress}
\end{table*}

\subsection{Combining SlimVLM with Token Pruning Methods}
Additionally, Our SlimVLM can seamlessly integrate with current token pruning methods. As shown in Table \ref{Combining_SlimVLM}, combining the performance-recovered SlimVLM pruned model with FastV results in a more lightweight model while maintaining excellent performance.

\renewcommand{\arraystretch}{1.2} 
\setlength\heavyrulewidth{0.25ex}
\begin{table*}[htbp!]
\centering
\small

\caption{Performance of pruned LLaVA-1.5-7B before and after integrating FastV token compression, with and without recovery.}

\resizebox{\textwidth}{!}{%
\begin{tabular}{cc|ccccccccc|c}
\toprule
  & FastV  & GQA  & MMB & MMB$^{\text{CN}}$ & MME & POPE & SQA$^{\text{I}}$ & VQA$^{\text{2}}$ & VQA$^{\text{T}}$ & VizWiz & Drop  \\ \cline{1-12}    
\multirow{2}{*}{w/o tune} & $\times$ & 58.00 & 56.44 & 45.88 & 1682.96 & 85.49 & 61.33 & 65.90 & 41.56 & 54.55 & - \\
 
& $\checkmark$  & 57.62 & 55.67 & 44.84 & 1706.18 & 85.18 & 61.08 & 65.68 & 42.11 & 51.11 & $\downarrow$ 0.99\%\\
\midrule
\multirow{2}{*}{w/ tune} & $\times$ & 59.67 & 58.99 & 48.26 & 1712.84 & 85.45 & 65.18 & 67.18 & 44.38 & 54.66 & - \\

& $\checkmark$ & 58.62 & 57.81 & 47.11 & 1708.18 & 85.18 & 64.04 & 66.68 & 43.17 & 53.73 & $\downarrow$ 1.52\% \\
\bottomrule
\end{tabular}
}
\label{Combining_SlimVLM}
\end{table*}

\subsection{Inference Efficiency}

Figure \ref{MMEGPU} shows the trade-off between performance and efficiency when applying SlimVLM to the LLaVA-1.5-7B model on the MME benchmark, under different pruning ratios—capturing changes in accuracy, inference speed, and GPU memory consumption.

\section{Visualization of adaptive visual token selection.}

\begin{figure}[htbp!]
  \centering
  \includegraphics[width=0.5\linewidth]{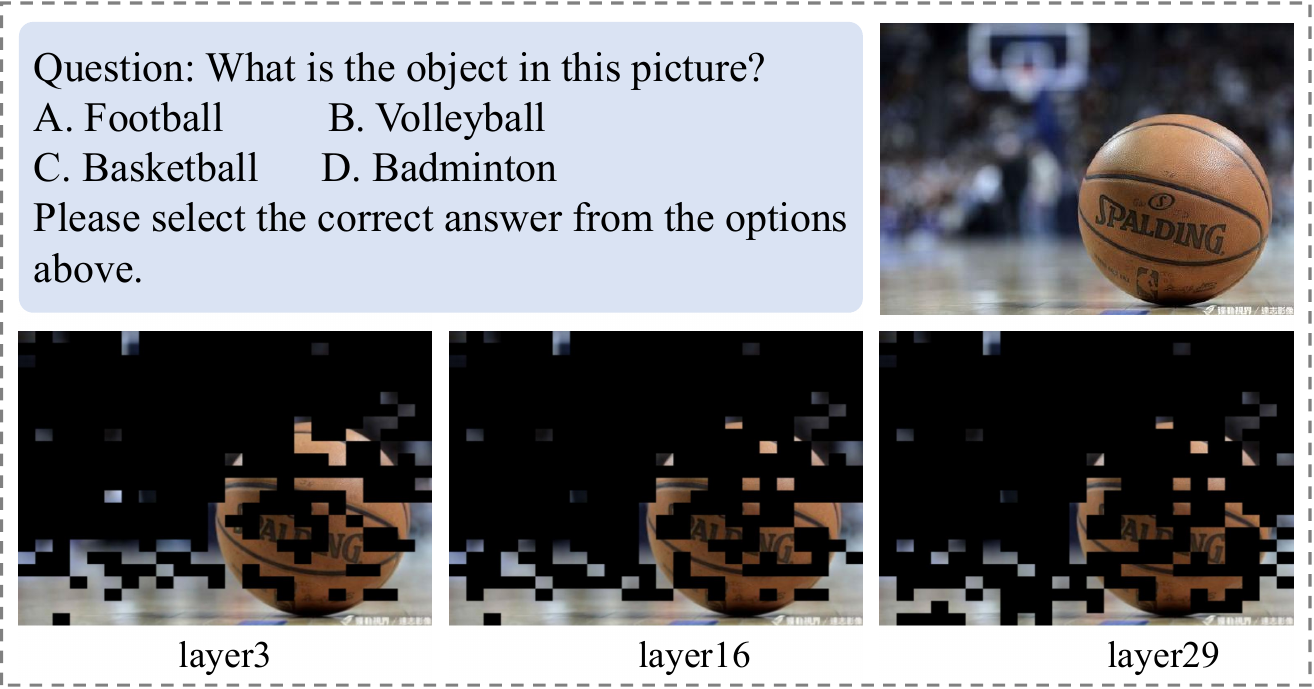}

   \caption{This figure illustrates the retained visual tokens at layers [3, 16, 29] for a sample from the MMBench dataset. It demonstrates that a significant portion of the discarded tokens originate from the background.}
   \label{token_select}
\end{figure}

Figure \ref{token_select} visualizes our token selection process. The method filters out most redundant tokens, and as the layers deepen, the model converges to a small number of core visual tokens highly relevant to the target.

\section{Limitations}
Although SlimVLM demonstrates strong pruning performance across different VLM architectures, it still has several limitations. First, due to limited computational resources, our method relies on a relatively small development set to estimate module importance and determine pruning ratios. While our experiments show that SlimVLM remains robust under different sampling sizes, the quality and diversity of the development samples may still affect the reliability of importance estimation, especially for domain-specific or highly specialized multimodal tasks. In addition, due to the substantial computational cost of pruning and evaluating large-scale VLMs, our experiments currently scale up only to LLaVA-1.5-13B. Further validation on larger-scale models, video-language models, and more diverse real-world deployment scenarios would help provide a more comprehensive assessment of the generality of our method.

\section{Visual Answer}
\begin{center}
  \includegraphics[width=\textwidth]{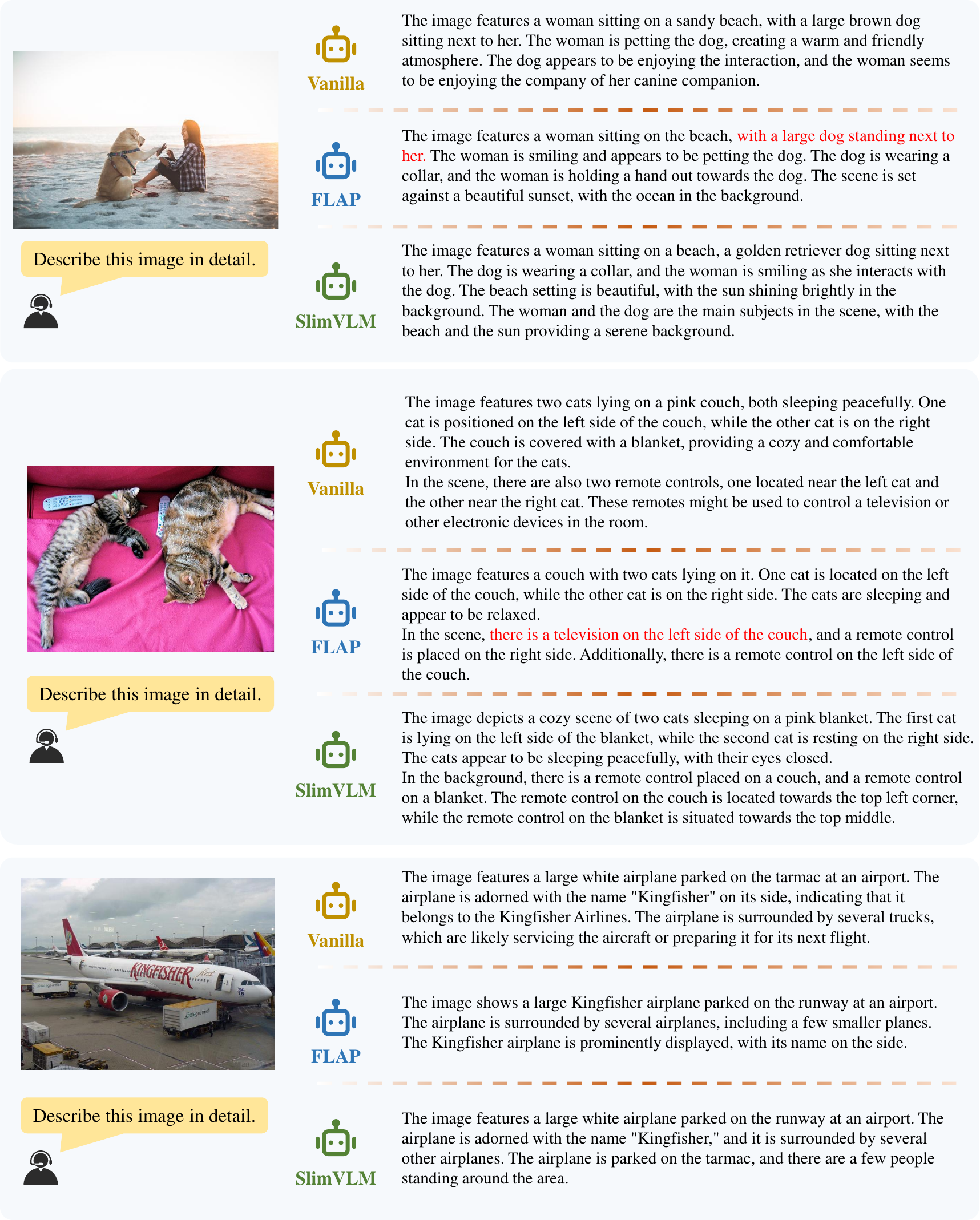}
  \captionof{figure}{Visual examples from LLaVA-1.5-7B under 20\% pruning ratio.}
  \label{...}
\end{center}

\end{document}